\documentclass[10pt,twocolumn,letterpaper]{article}

\usepackage{cvpr}
\usepackage{times}
\usepackage{epsfig}
\usepackage{graphicx}
\usepackage{subcaption}
\usepackage{amsmath}
\usepackage{amssymb}
\usepackage{bbold}
\usepackage{soul}
\usepackage{authblk}
\usepackage{float}

\usepackage[breaklinks=true,bookmarks=false]{hyperref}

\cvprfinalcopy 

\def\cvprPaperID{****} 

\begin{document}

\title{Detection and Localization of Image Forgeries using Resampling Features and Deep Learning}




\author[1]{Jason Bunk}
\author[2]{Jawadul H. Bappy}
\author[1]{Tajuddin Manhar Mohammed}
\author[1]{Lakshmanan Nataraj}
\author[3]{Arjuna Flenner}
\author[1,4]{B.S. Manjunath}
\author[1,4]{Shivkumar Chandrasekaran}
\author[2]{Amit K. Roy-Chowdhury}
\author[3]{Lawrence Peterson}
\affil[1]{Mayachitra Inc., Santa Barbara, California , USA}
\affil[2]{Department of Electrical and Computer Engineering, University of California, Riverside, USA}
\affil[3]{Naval Air Warfare Center Weapons Division, China Lake, California, USA}
\affil[4]{Department of Electrical and Computer Engineering, University of California, Santa Barbara, USA}

\date{\vspace{-5ex}}

\maketitle

\begin{abstract}

Resampling is an important signature of manipulated images. In this paper, we propose two methods to detect and localize image manipulations based on a combination of resampling features and deep learning.
In the first method, the Radon transform of resampling features are computed on overlapping image patches. Deep learning classifiers and a Gaussian conditional random field model are then used to create a heatmap. Tampered regions are located using a Random Walker segmentation method.
In the second method, resampling features computed on overlapping image patches are passed through a Long short-term memory (LSTM) based network for classification and localization.
We compare the performance of detection/localization of both these methods. Our experimental results show that both techniques are effective in detecting and localizing digital image forgeries.
  
\end{abstract}

\section{Introduction}

The number of digital images has grown exponentially with the advent of new cameras, smartphones and tablets. 
Social media such as Facebook, Instagram and Twitter have further contributed to their distribution. 
Similarly, tools for digitally manipulating these images have evolved significantly, and software such as Photoshop, Gimp and smartphone apps such as Snapseed, Pixlr make it very trivial for users to easily manipulate images.
Several methods have been proposed to detect digital image manipulations based on artifacts from resampling, color filter array, lighting, camera forensics, JPEG compression, and many more. 
In this paper, we consider artifacts that arise due to resampling which is common when creating digital manipulations such as scaling, rotation or splicing.
We propose two methods to detect and segment image forgeries.
In the first method, we present an end-to-end system to detect and localize these digital manipulations based on Radon transform and Deep Learning.
In the second, we use a combination of resampling features based on probability-maps (p-maps) and Long short-term memory (LSTM) based modeling to classify tampered patches.


\begin{figure}[H]
\centering
\includegraphics[width=8 cm]{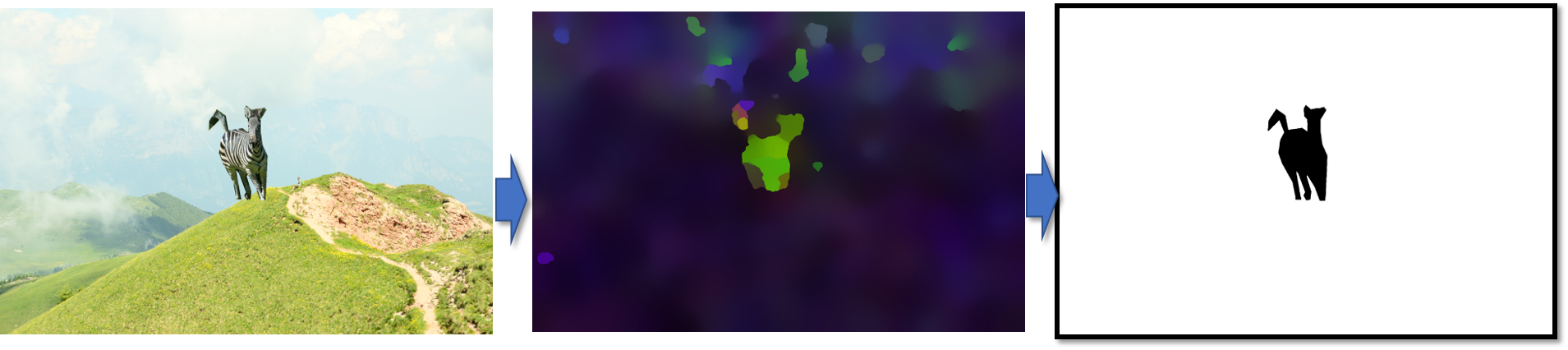}
\vspace{-5pt}
\caption{Illustration of our system to detect and localize digital manipulations. The manipulated image on the left is processed through a bank of resampling detectors to create a heat-map (middle), which is then used by a classifier to detect/localize the modified regions (right)} 
\label{fig:illus}
\vspace{-10pt} 
\end{figure}

The rest of the paper is organized as follows. 
In Section~\ref{sec:relwork}, we detail some of the related work in the field of Digital Image Forensics.
In Section~\ref{sec:frame}, we describe our end-to-end system to detect and localize manipulations.
Section~\ref{sec:exps} details the experimental results while the conclusion and future work is presented in Section~\ref{sec:conc}.


\begin{figure*}[ht]
\centering
\includegraphics[width=\linewidth]{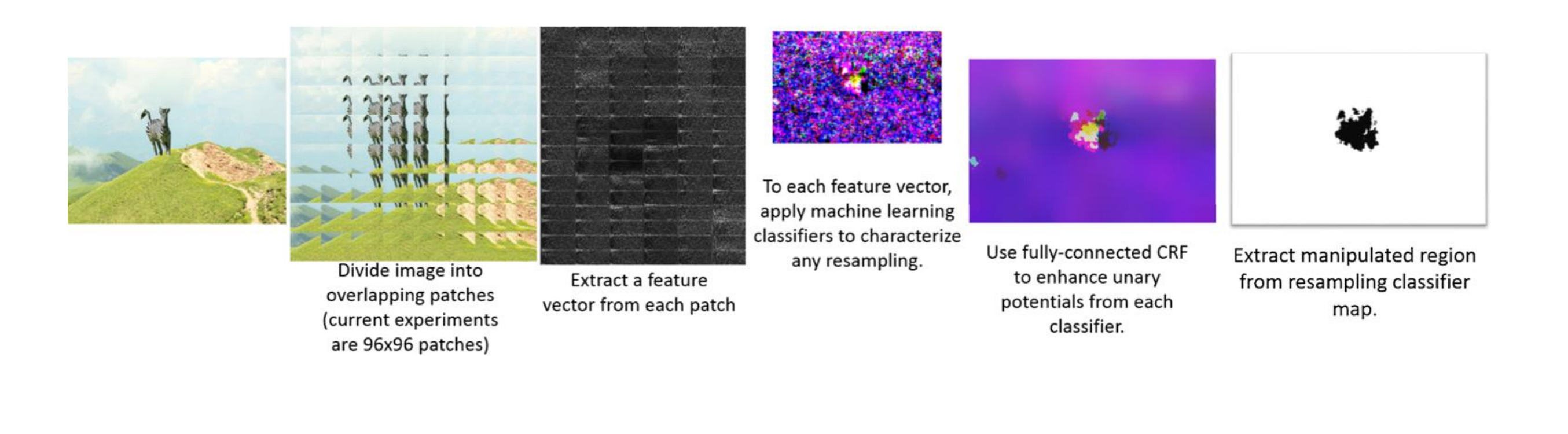}
\vspace{-25pt}
\caption{An end-to-end framework to detect and localize digital manipulations.} 
\label{fig:blkschm}
\vspace{-10pt} 
\end{figure*}

\section{Related Work}
\label{sec:relwork}

The field of image forensics comprises diverse areas to detect a manipulated image which includes
resampling detection, detection of copy-move, splicing and object removal, JPEG artifacts, machine learning and deep learning techniques.
We will briefly discuss some of them below.

In the past decade several techniques have been proposed to detect resampling in digital images~\cite{popescu-farid-resampling, prasad2006resampling, second-diff-method, babak-radon, kirchner-local, feng2012normalized, ryu2014estimation}. 
In most cases, it is assumed to be done using linear or cubic interpolation.
In~\cite{popescu-farid-resampling}, the authors discuss how resampling introduces specific statistical correlations and show that they can be automatically detected using an Expectation-Maximization (EM) algorithm.
The algorithm estimates the periodic correlations (after first detecting whether such a correlation exists) among the interpolated pixels -
the specific type of the correlation indicates the exact form of the resampling.
However, the EM-based method is very susceptible to JPEG attacks - especially when the JPEG quality factor (QF) is 95 or lower.
The reason is that the periodic JPEG blocking artifacts interfere with the periodic patterns introduced by resampling.
For images that are scaled using linear/cubic interpolation, an algorithm was proposed by analyzing the variance of the second difference of interpolated signals~\cite{second-diff-method}. 
Although this method can detect only up-scaling, it is very robust against JPEG compression and detection is
possible even at very low QFs. (Downscaled images can be detected up to a certain extent but not as robustly as upscaled images.) 
This method was further improved to tackle other forms of resampling using a Radon transform and derivative filter based approach~\cite{babak-radon}. 
In this paper, we utilize the Radon transform based method and build a feature to detect manipulated regions (see Sec.~\ref{sec:deep-resamp}).
In~\cite{ kirchner-local}, the author showed a simpler method to improve~\cite{popescu-farid-resampling} by using a linear predictor residue instead of the computationally expensive EM algorithm.
We combine this method with deep learning based models to detect tampered blocks (see Sec.~\ref{LSTM-Conv}).
In~\cite{ryu2014estimation}, the authors exploit  periodic properties of interpolation by the second-derivative of the transformed image for detecting image manipulation.
To detect resampling on JPEG compressed images, the authors added noise before passing the image through the resampling detector and showed that adding noise aids in detecting resampling~\cite{Nataraj10-345,Nataraj09-331}. 
In~\cite{feng2011energy,feng2012normalized}, a feature is derived from the normalized energy density and then SVM is used to robustly detect resampled images. 
Some recent approaches~\cite{golestaneh2014algorithm, kwon2015efficient} have been proposed to reduce the JPEG artifact left by compression.

Many methods have been proposed to detect copy-move~\cite{li2015segmentation, cozzolino2015efficient}, splicing~\cite{guillemot2014image,muhammad2014image}, seam carving~\cite{sarkar2009detection,liu2015improved} and inpainting based object removal~\cite{wu2008detection,liang2015efficient}. 
Several approaches exploit JPEG blocking artifacts to detect tampered regions~\cite{farid2009exposing,lin2009fast,luo2010jpeg,bianchi2011improved}. 
In computer vision, deep learning shows outstanding performance in different visual recognition tasks such as image classification~\cite{zhou2014learning}, and semantic segmentation~\cite{long2015fully}. 
In~\cite{long2015fully}, two fully convolution layers have been exploited to segment different objects in an image. The segmentation task has been further improved in \cite{zheng2015conditional, badrinarayanan2017segnet}. 
These models extract hierarchical features to represent the visual concept, which is useful in object segmentation. Since, the manipulation does not exhibit any visual change with respect to genuine images, these models do not perform well in segmenting manipulated regions. 

Recent efforts in detecting manipulations exploit deep learning based model in~\cite{bayar2016deep, bayar2017design,rao2016deep}.
These include detection of generic manipulations~\cite{bayar2016deep, bayar2017design}, resampling~\cite{bayar2017resampling}, splicing~\cite{rao2016deep} and bootleg~\cite{buccoli2014unsupervised}.
The authors tested existing CNN network for steganalysis~\cite{qian2015deep}. 
In~\cite{qian2015deep}, the authors propose Gaussian-Neuron CNN (GNCNN) for steganalysis.
A deep learning approach to identify facial retouching was proposed in~\cite{bharati2016detecting}.
In~\cite{zhang2016image}, image region forgery detection has been performed using stacked auto-encoder model. 
In\cite{bayar2016deep},  a new form convolutional layer is proposed to learn the manipulated features  from an image. Unlike most of the deep learning based image tampering detection methods which use convolution layers, we present an unique network exploiting convolution layers along with LSTM network.

\section{Detection and Localization}
\label{sec:frame}

In this section, we describe  our end-to-end system to detect and localize manipulations in digital images.
Fig.~\ref{fig:blkschm} shows the block schematic of our system. 
The various steps in framework are as follows:
We start by extracting small, overlapping patches. As the first step in the processing pipeline, we compute features on each patch. These are used to characterize any resampling applied to the patch. This produces a multi-channel heatmap, one channel per resampling characteristic, at every point at which the patches were extracted. By densely extracting overlapping patches (stride of 8), we can interpret a correspondence between a pixel from the original image and the point in the heatmap representing the patch centered at that pixel. As final steps, we postprocess this heatmap and use it to produce an image-level detection score and binarized localization map. 
The manipulated regions are then extracted using image segmentation methods based on Otsu's thresholding and Random Walk segmentation.

There is a tradeoff in selecting the patch size: resampling is more detectable in larger patch sizes because the resampling signal has more repetitions, but small manipulated regions will not be localized that well. We choose 64x64 as a small size that we can detect reasonably well, as will be discussed in section 3.1. We extract patches with a stride of 8 for computational efficiency and to have a consistent response w.r.t. 8x8 JPEG blocks. The features in the first step start with the absolute value of a 3x3 Laplacian filter, which produces an image of the magnitude of linear predictive error. To look for periodic correlations in the linear predictor error, we apply the Radon transform to accumulate errors along various angles of projection, and then take the FFT to find the periodicity. This was proposed in~\cite{babak-radon}. We performed some additional experiments on this feature extraction stage to try to explore any potential variations that could improve these features. One improvement we found was to take the square root of the magnitude after Laplacian filtering. We also compared with using the residual from a 3x3 Median filter and found that the 3x3 Laplacian produces 4.6\% better results (mean AUC over 6 binary classification tasks).

The second step is to characterize any resampling detected in the patch. We train a set of six binary classifiers that check for different types of resampling. The six resampling characteristics are: JPEG quality thresholded above or below 85, upsampling, downsampling, rotation clockwise, rotation counterclockwise, and shearing (in an affine transformation matrix). To train a model for each task, we build a dataset of about 100,000 patches extracted from about 8,000 images from two publicly available datasets of raw uncompressed images, UCID~\cite{schaefer2003ucid} and RAISE~\cite{dang2015raise} datasets. 
Some of the patches are transformed with a set of randomly generated parameters, like multiple JPEG compressions and affine transformations, but one half of the dataset must include the transformation specified, and the other half must not.
The classifiers are not mutually exclusive, and are trained individually. 
The best performing classifier we found for this task was an artificial neural network with two hidden layers. Using cross-validation this beat a Bayesian quadratic classifier by 6.5\% on mean AUC (0.82 vs 0.77 mean AUC averaged over the six classifiers).
The filtering in the third step uses bilateral filtering, which has been shown to improve region detection accuracy in other domains like semantic segmentation when fusing adjacent noisy local patch-based classifiers ~\cite{koltun2011efficient}.
The fourth and final step is to obtain a mask showing manipulated regions from the filtered resampling heatmaps using proposed selective segmentation model.





\subsection{Deep Neural Networks for Resampling Detection}
\label{sec:deep-resamp}

\begin{figure}[h]
\centering
\includegraphics[width=\columnwidth]{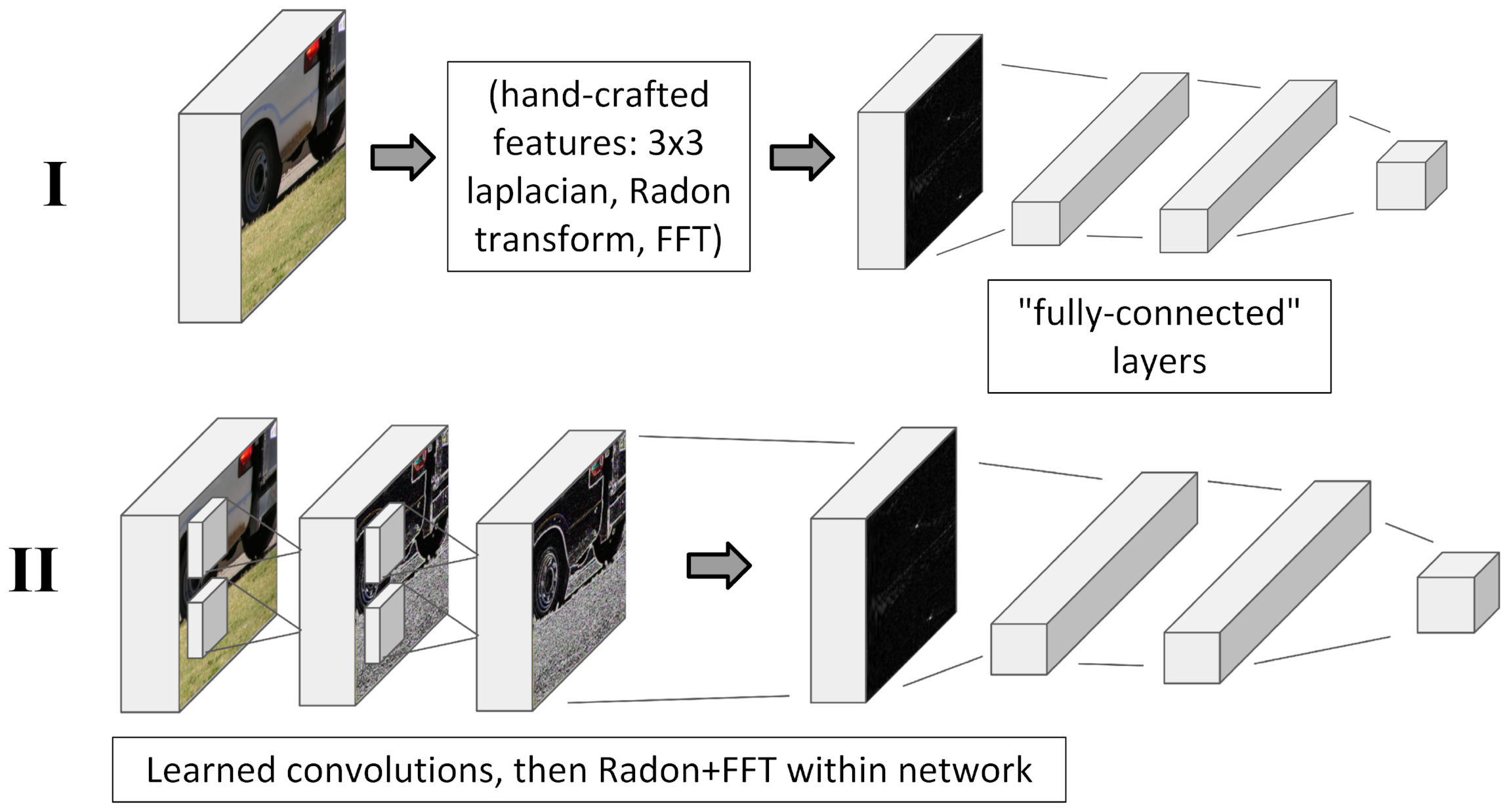}
\vspace{-15pt}
\caption{Deep Neural networks for detecting resampling in small patches.} 
\label{fig:resamp-nets}
\vspace{-10pt} 
\end{figure}

As described in the previous section, we found that an artificial neural network with two hidden layers performed best as a binary classifier characterizing resampling in a patch for step 2 in our detection pipeline. We also explored using a newly proposed convolutional layer~\cite{bayar2016deep} as the first layer of an end-to-end patch classification network to learn to extract resampling features itself, rather than using our hand-crafted resampling features described before as step 1 of our pipeline. These two neural network architectures are graphed visually in Fig.~\ref{fig:resamp-nets}.

We experimented with using an FFT within the network of model II, made possible by Tensorflow \cite{abadi2016tensorflow}, and inspired by model I. Compared to a model that did not include the FFT, we found that the inclusion of the FFT stabilized training, reducing noise in the training loss, and caused the model to converge faster and to a slightly higher (1\%) final score.

\begin{figure*}[ht]
\centering
\begin{minipage}{.48\textwidth}
    \centering
    \captionsetup{width=.9\linewidth}
    \includegraphics[width=0.95\linewidth]{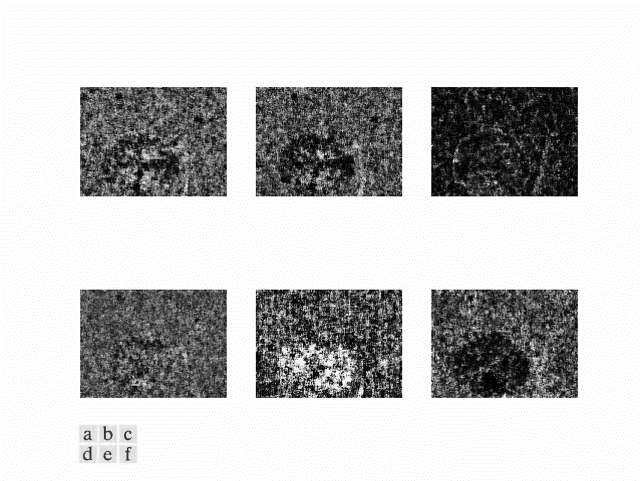}
    \vspace{-10pt}
    \caption{(a)-(f) Outputs from the neural network based resampling detector of a manipulated image} \
    \label{fig:unfilt-clf-op}
\end{minipage}%
\begin{minipage}{.48\textwidth}
    \centering
    \captionsetup{width=.9\linewidth}
    \includegraphics[width=0.8\linewidth]{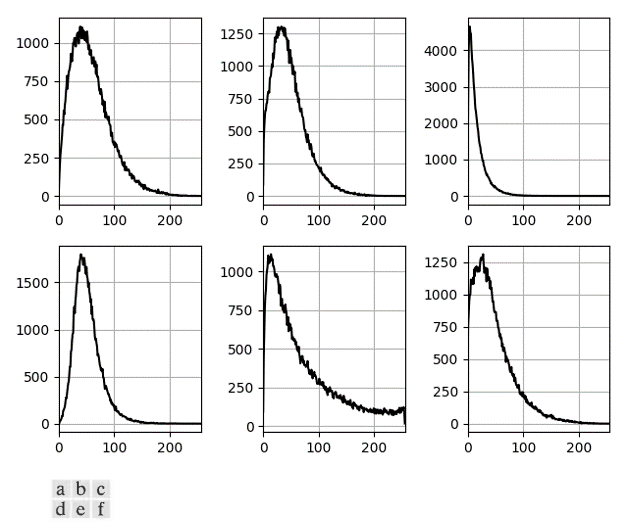}
    \vspace{-10pt}
    \caption{(a)-(f) Corresponding histograms of images in Fig.~\ref{fig:unfilt-clf-op}} 
    \label{fig:unfilt-clf-hist}
\end{minipage}%
\end{figure*}

We found that the first architecture (I) using hand-crafted features performed slightly better at detecting rotation, shearing, and downsampling, but the second architecture (II) performed significantly better at detecting upsampling and differences in JPEG compression quality factors. The best ROC AUC scores of the methods are listed in the table below, and the model from which the best score was taken is listed in the rightmost column. Model II performed 1\% to 3\% worse on the four lower rows, which was a small but consistent difference that could be due to optimization difficulties. Model I's best ROC AUC scores for the first two rows, JPG quality and rescaling up, were 0.87 and 0.89 respectively.

\vspace{2pt}
\begin{table}[htb]
\centering
\begin{tabular}{ccc}
\cline{2-3}
\multicolumn{1}{c|}{}   & \multicolumn{1}{c|}{AUC} & \multicolumn{1}{c|}{Model}  \\ \cline{1-3}
\multicolumn{1}{|c|}{JPG quality} & \multicolumn{1}{c|}{0.93} & \multicolumn{1}{c|}{II.} \\ \cline{1-3}
\multicolumn{1}{|c|}{Rescale up} & \multicolumn{1}{c|}{0.92} & \multicolumn{1}{c|}{II.}  \\ \cline{1-3}
\multicolumn{1}{|c|}{Rescale down} & \multicolumn{1}{c|}{0.77} & \multicolumn{1}{c|}{I.}  \\ \cline{1-3}
\multicolumn{1}{|c|}{Rotate CW} & \multicolumn{1}{c|}{0.81} & \multicolumn{1}{c|}{I.}  \\ \cline{1-3}
\multicolumn{1}{|c|}{Rotate CCW} & \multicolumn{1}{c|}{0.81} & \multicolumn{1}{c|}{I.}  \\ \cline{1-3}
\multicolumn{1}{|c|}{Shearing} & \multicolumn{1}{c|}{0.76} & \multicolumn{1}{c|}{I.}  \\ \cline{1-3}
\end{tabular}
\caption{Experiments on patch classification}
\end{table}


The scores on patch resampling classification depend on the patch size, particularly the JPEG quality level detection and scaling up. We compared against another state-of-the-art deep learning model~\cite{bayar2017robustness}, which was trained on larger 256x256 patch sizes. 
The evaluation metric they use is accuracy, while we use AUC which is a more robust evaluation metric. Averaged over JPEG quality factors from 60 to 100, the authors report 96.8\% accuracy for detecting upscaling. In our experiments, we report the trend in AUC of 0.922 on 64x64 patches and AUC 0.950 on 128x128 patches. Following the trend in increasing accuracy with increasing patch size, we are at a similar level of accuracy. 
In addition, a fair and direct comparison would require tuning each method to use the best patch size and on the same dataset.






\subsection{Mask Filtering and Segmentation}

Here we describe how we filter the heatmaps and segment the manipulated regions.



\begin{figure}[b]
\centering
\includegraphics[width=\linewidth]{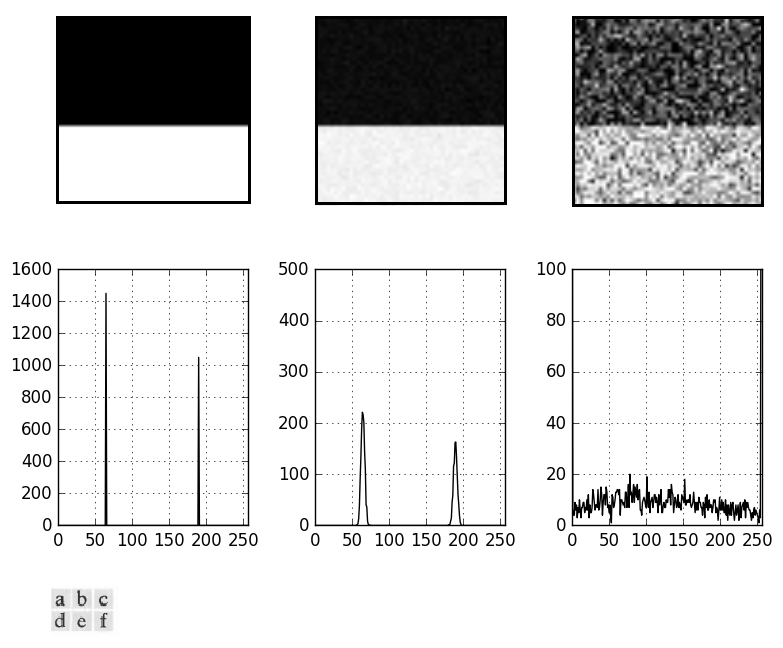}
\vspace{-15pt}
\caption{(a) Noiseless 8-bit image. (b) Image with additive Gaussian noise of mean 0 and variance 0.0001. (c) Image with additive Gaussian noise of mean 0 and variance 0.06. (d)-(f) Corresponding histograms} 
\label{fig:img-noise}
\end{figure}

The noisy feature maps (Fig.~\ref{fig:unfilt-clf-op}) can be filtered to enhance segmentation. 
As shown in the Fig.~\ref{fig:unfilt-clf-hist}, the gray level histograms of these maps have “nice” properties with respect to their distribution (unimodal).
This distribution can be compared to that of a special case of adding excessive gaussian-like noise to the binary images. 
As seen in Fig.~\ref{fig:img-noise}, the third image from the left that was obtained after adding excessive white gaussian noise and has a similar histogram distribution to that of our feature maps. 
The goal is to obtain a bimodal-like distribution for the histogram of the feature maps after filtering (similar to that of the second image in Fig.~\ref{fig:img-noise}) that will be useful for the segmentation task. 

\begin{figure*}[ht]
\centering
\begin{minipage}{.5\textwidth}
    \centering
    \includegraphics[width=8.2 cm]{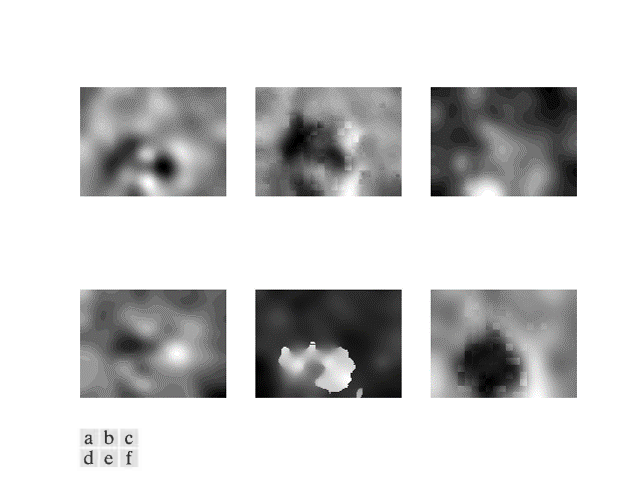}
    \vspace{-8pt}
    \caption{(a)-(f) Bilateral filtered heatmaps} 
    \label{fig:filt-clf-op}
\end{minipage}%
\begin{minipage}{.5\textwidth}
    \centering
    \captionsetup{width=.9\linewidth}
    \includegraphics[width=8 cm]{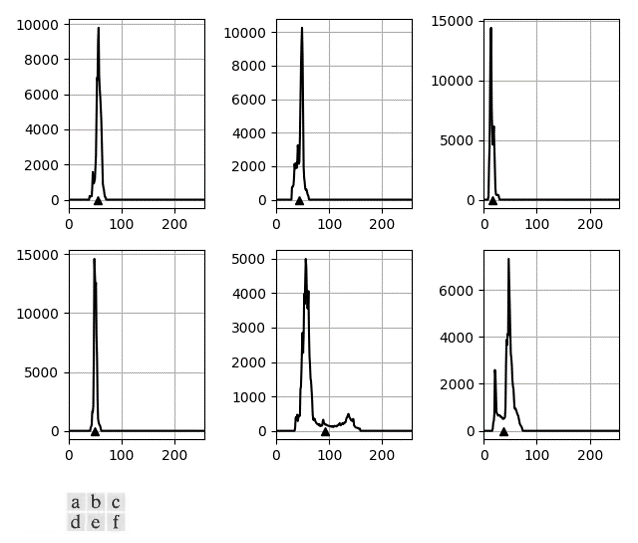}
    \vspace{-1pt}
    \caption{(a)-(f) Corresponding histograms of images in Fig.~\ref{fig:filt-clf-op} with Otsu thresholds marked with $``\blacktriangle"$}
    \label{fig:filt-clf-hist}
\end{minipage}%
\end{figure*}

We used edge-preserving bilateral filters as inspired by their successful application in the semantic segmentation field~\cite{koltun2011efficient,zheng2015conditional}, a mathematically analogous task of filtering a set of noisy classifier outputs predicted on overlapping patches. The bilateral filter parameters were empirically determined from the noise distribution of various feature maps. 
As shown in the Fig.~\ref{fig:filt-clf-hist}, only a subset of these six feature maps have a bimodal-like distribution in their histograms after filtering. Therefore, the ones which does not have such a distribution are not used for segmentation.


The feature maps that are used for the segmentation is determined by Otsu method of thresholding image histograms~\cite{otsu1975threshold} where it can detect bimodal distributions  based on the fact that this threshold minimizes the intra-class variance and maximizes the inter-class variances of two predicted classes. As seen in the Fig.~\ref{fig:filt-clf-hist}, the Otsu threshold is either a value close to the mean of the distribution or it separates the two classes of a bimodal distribution.

A random-walker segmentation~\cite{grady2007randomwalk} is then applied to each of these maps to diffuse the corresponding pixels in between the two modes to either of the classes (class-1 being the pixel values less than the first mode and class-2 being the pixel values greater than the second mode). This probabilistic based method is useful as it locates the weak/missing boundaries and is even robust in noisier conditions. Finally, a bitwise-OR operation is used to add all these binary images to obtain a final mask image. 


\subsubsection{Localization based confidence score metric}

Bilateral filtering is implemented using ArrayFire package in Python which uses GPU's for faster computations by making some approximations. Therefore, a gray-scale mask is generated instead of a binary mask (by considering the mean of all the binary masks for many iterations). The confidence score is generated by adding the normalized non-zero pixel values (normalized pixel value of 1 corresponds to a black or manipulated region) and averaging them over the number of non-zero pixels in the gray-scale mask image.


\subsection{ Patch Classification Framework}
\label{LSTM-Conv}
Now-a-days authentication of image tampering is a very difficult problem due to the close resemblance between the manipulated and genuine images.
However, most of the image manipulations have a common characteristic, which shows discriminative features in the boundary shared between manipulated and non-manipulated regions. 
In this paper, we present a network which exploits resampling features along with  long-short term memory (LSTM) cells in order to localize the manipulated regions.

\begin{figure}[ht]
\centering
\includegraphics[width=8 cm]{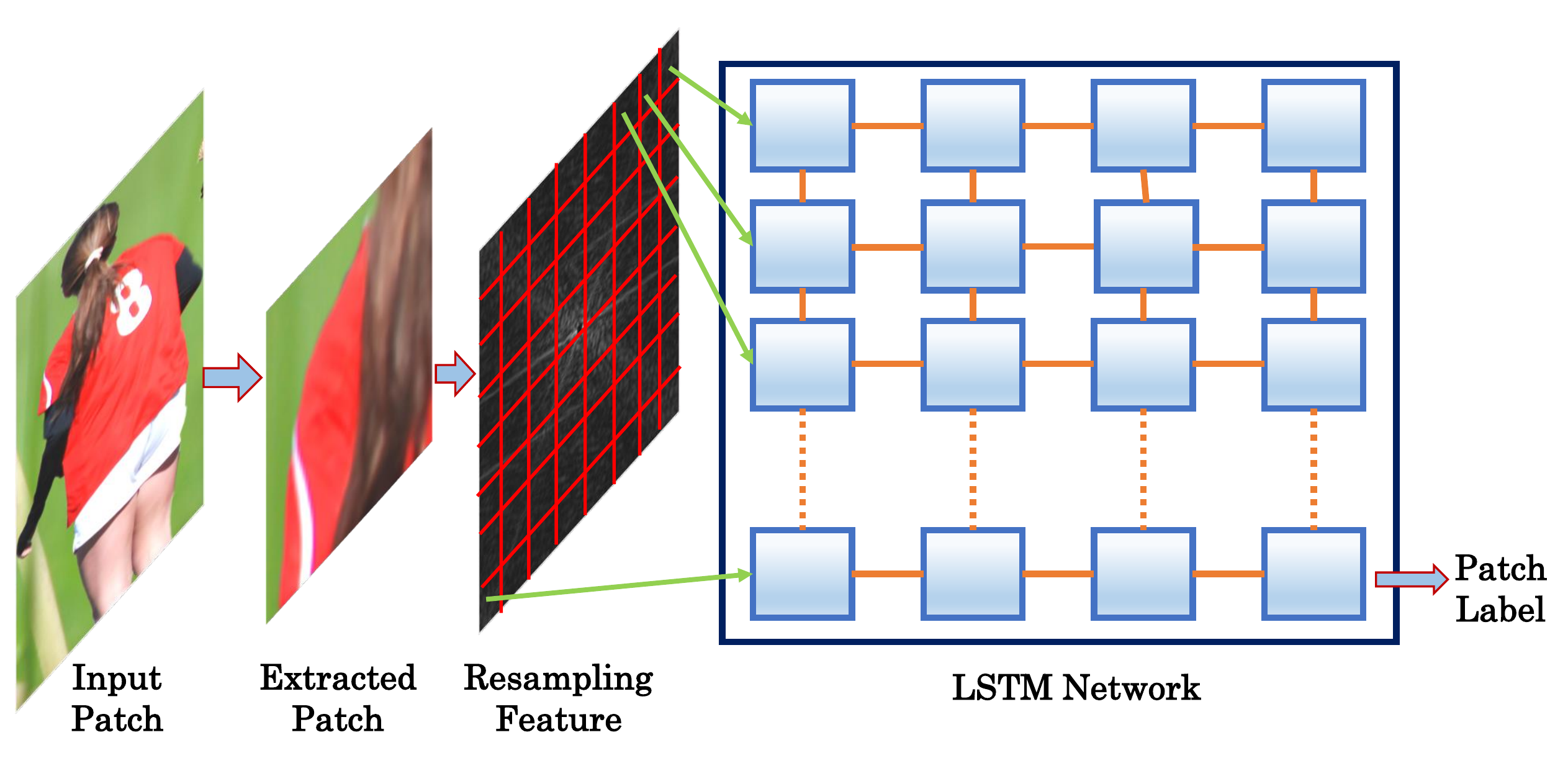}
\vspace{-15pt}
\caption{LSTM Framework for Patch Classification.} 
\label{patch_framework}
\end{figure}

\subsubsection{Resampling Features}

In~\cite{popescu-farid-resampling}, the authors proposed the seminal method to detect resampling in digital images.
We will provide a brief introduction to their approach. 
The idea is that resampling introduces periodic correlations among pixels due to interpolation.
To detect these correlations, they use a linear model in which each pixel is assumed to belong to two
classes: a resampled class and a non resampled class , each with equal probability. 
The conditional probability for a pixel belonging to the resampled class is assumed to be Gaussian while the conditional probability for a pixel belonging to the other class is assumed to be uniform.
To simultaneously estimate a pixel's probability of being a linear combination with its neighboring pixels and the unknown weights of the combination, an Expectation Maximization (EM) algorithm is
used.
In the Expectation Step, the probability of a pixel belonging to the resampled class is calculated.
This is used in the Maximization step to estimate the weights.
The stopping condition is enforced when the difference in weights between two consecutive iterations is very small.
At this stage, the matrix (same dimensions as image) of probability values obtained in the Expectation step for every pixel of the image is called the ``Probability map (p-map)".
For a resampled image this p-map is periodic and peaks in the 2D Fourier spectrum of the p-map indicate resampling. 
In the p-map, a probability value close to 1 indicates that a pixel is resampled. 
However, the EM algorithm based p-map is computationally expensive. 
In this paper, we use a simpler method to compute the p-map that first filters the image using a linear filter and then computes the residual image on which the p-map was calculated~\cite{kirchner-local}.

\subsubsection{Long-Short Term Memory (LSTM) Network}
\label{LSTM}
Manipulation distorts the natural statistics of an image, especially in the boundary region. 
In this paper, we utilize the LSTM network to learn the correlation between blocks of resampling features as shown in Fig.~\ref{patch_framework}. 
In order to utilize LSTM, we first divide the 2D resampling feature map into blocks. We split into $ 8\times 8 $ blocks, where each block has size $ 8\times 8 $ (total $ 64 $ pixels). Now, we learn the long distance block dependency feeding each block to each cell of the LSTM in a sequential manner. 
The LSTM cells correlate neighboring blocks with current block. In this work, we utilize $ 3 $ stacked layers, and at each layer, $ 64 $ cells are used. 
In the last layer, each cell provides $256$-d feature vector which is fed into softmax classifier. The key insight of using LSTM is to learn the boundary transformation between different blocks, which provides discriminative features between manipulated and non-manipulated patch. 
In the following, we will briefly discuss about the parameters of a LSTM cell.


The cell is an important entity to build an LSTM network. The information flow in the cells is controlled by gates. There are mainly three gates that link a cell to neighboring cell, and they are 
 (1)  input gate, (2) forget gate, and (3)  output gate. 
 Let us denote cell state and output state as $C_t$ and  $z_t$ for current cell $t$. 
Cell state and output states are controlled by these gates. 
Each gate has a value ranging from zero to one, activated by a sigmoid function. These gates  actually make decisions about  how much information should be passed through. Higher value implies the flow of more information.
Each cell produces new
candidate cell state $\bar{\mathcal{C}}_t $. Using the previous cell state $ \mathcal{C}_{t-1} $ and $\bar{C}_t$, we can write the updated cell state $ \mathcal{C}_t $  as  
\begin{equation}
\label{cell_state}
\mathcal{C}_t=f_t \circ \mathcal{C}_{t-1}+i_t \circ \bar{\mathcal{C}}_t
\end{equation}
Here, $ \circ $ denotes the $pointwise$ multiplication. 
Finally, we obtain the output of the current cell $ h_t $, which can be represented as 
\begin{equation}
\label{out_state}
h_t = o_t\circ tanh(\mathcal{C}_t)
\end{equation}
In Eqns.~\ref{cell_state} and ~\ref{out_state}, $i,f,o $ represent input, forget and output gates. 


\subsubsection{Soft-max Layer}
In the proposed network, we have a softmax layer for patch classification task. For this task, the labels are predicted  at the final layer of LSTM network.  
Given a patch of an image, 
we obtain the features which are used  to predict the manipulated class (either patch label or pixel class) using \textit{softmax} function. 
Let $ \mathcal{W} $ be the parameter associated with feature. 
Then, $\mathcal{F}$,
the softmax function can be written as 

\begin{equation}
\label{classification}
P(\mathcal{Y}_k)=\frac{\exp^{(\mathcal{W})^T\mathcal{F}}}{\sum_{k=1}^{N_c}\exp^{(\mathcal{W}^k)^T\mathcal{F}}}
\end{equation}
Here,  $ N_c $ is the number of classes ($ \in \mathcal{R}^{2\times 1} $) - manipulated vs non-manipulated. $ \mathcal{W}_{L}^k $ implies  the weight vector associated to the class $ k $. 
Using Eqn.~\ref{classification}, we compute the probability distribution over various classes. 
Now, we can predict labels by maximizing $ P(\mathcal{Y}_k) $ with respect to $ k $. The predicted label can be obtained by $ \hat{\mathcal{Y}}=\arg\underset{k}\max \ \ P(\mathcal{Y}_k) $.

\subsubsection{Training the Network} 

In this paper, we perform patch classification whether it is manipulated or not. We  compute the cross entropy loss of this task  given ground-truth patch labels.  

\paragraph{Patch Classification.}
Patch labels are predicted at the end of the LSTM network as shown in Fig.~\ref{patch_framework}. Let us denote $\theta_p=[\theta_1, \mathcal{W}]$, which is a weight vector associated with patch classifiation. Here, $\theta_1$ contains the parameters of first two convlution layers and LSTM layers. $\mathcal{W}$ is the parameter at the softmax layer of patch classification. Now, we can compute the cross entropy loss for patch classification as follows. 
\begin{equation}
\label{loss_p}
\mathcal{L}_p(\theta_p)=-\frac{1}{M_p}\sum_{j=1}^{M_p}\sum_{k=1}^{N_p}\mathbb{1}(\mathcal{Y}^j=k)\log (\mathcal{Y}^j=k|x^j;\theta_p)
\vspace{-.5mm}
\end{equation}
Here, $ \mathbb{1}(.) $ is the \textit{indicator function}, which equals to $ 1 $ if $ j=k $, otherwise it equals $ 0 $. 
$ \mathcal{Y}^j $ and $ x^j $ imply the patch label (manipulated or non-manipulated) and the feature of the sample $ j $.
 $ M_p $ is the number of patches.

\begin{figure*}[ht]
\centering
\includegraphics[height=0.45\linewidth]{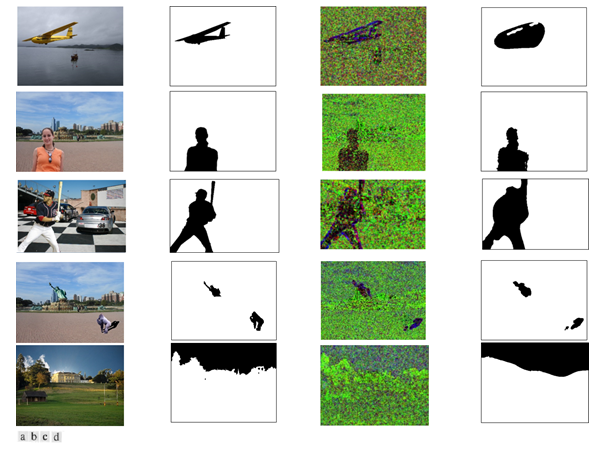}
\vspace{-5pt}
\caption[Caption for LOF]{(a) Manipulated Image (b) Ground truth (c) Resampling detector output visualizing 3 out of 6 classifiers. Each color represents a different type of manipulation. (d) Predicted mask} 
\label{fig:res}
\vspace{-10pt} 
\end{figure*}

\section{Experimental Results}
\label{sec:exps}

\subsection{Results on Radon transformed and Deep Learning based detection and localization}

In this section, we demonstrate our experimental results for two tasks-(1) identification of tampered patch, and (2) segmentation of manipulated regions given a patch.
We evaluate the proposed model on NIST Nimble 2016 dataset~\cite{2016-nimble-dataset} for the Media Forensics challenge. 
This dataset includes mainly three types of manipulation: (a) copy-clone, (b) removal, and (c) splicing. 
The images are tampered in a sophisticated way to beat current state-of-the-art detection techniques.
The results in Fig.~\ref{fig:res} show the efficacy of the proposed model for different images in Nimble 2016 dataset.


\subsection{Results on LSTM based patch classifier}

\begin{figure}[H]
\centering
\includegraphics[width=0.7\linewidth]{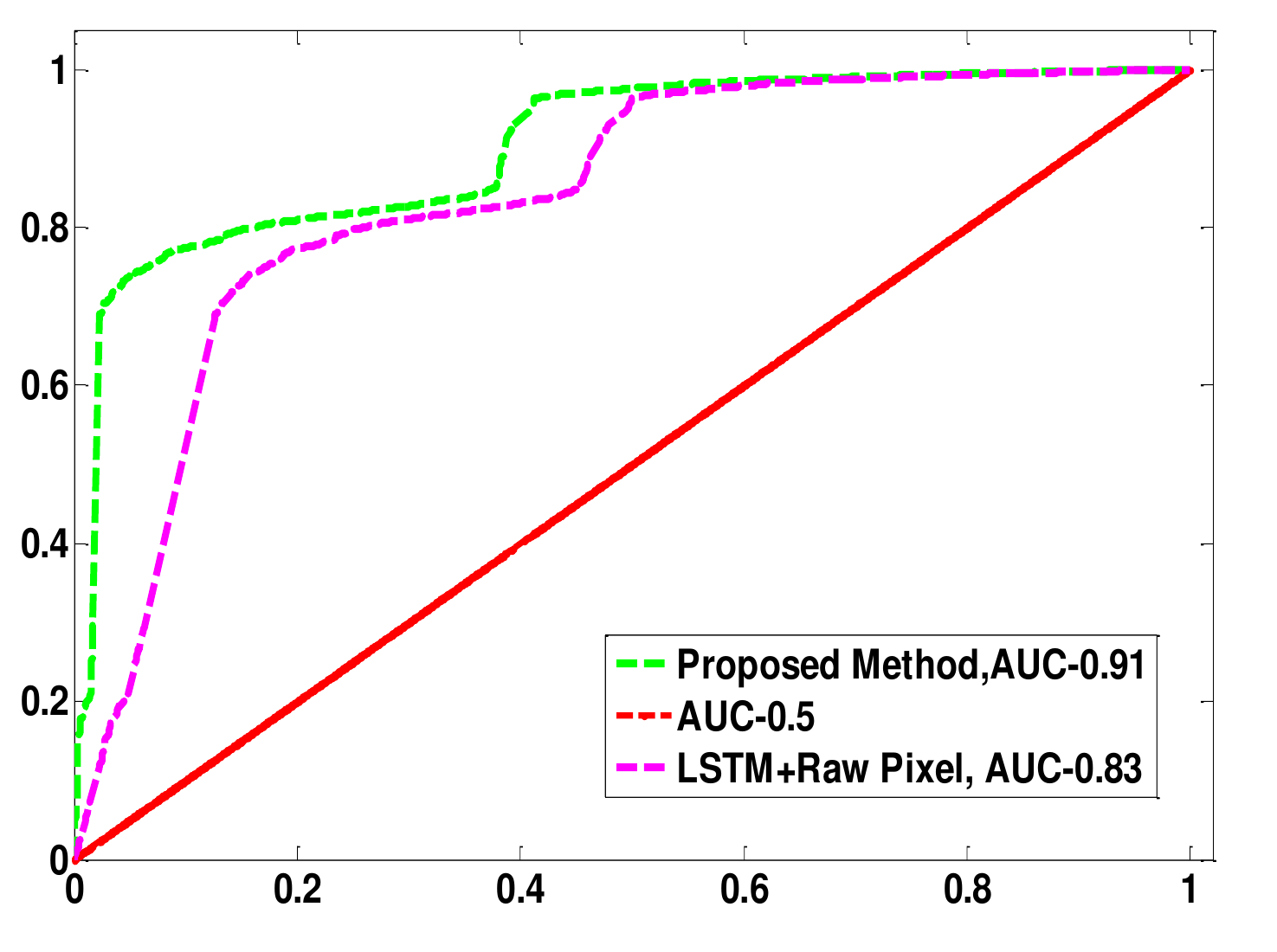}
\caption{ROC Curve comparing LSTM based method on Pixels and Resampling Detection features} 
\label{roc_curve}
\end{figure}

Here we again use the Nimble 2016 dataset~\cite{2016-nimble-dataset} to test the efficacy of LSTM based approach. 
We use ROC curves to evaluate our method. 
The curve is created by plotting the true positive rate (TPR) against the false positive rate (FPR) at various threshold. 
Fig.~\ref{roc_curve} shows ROC curve of the proposed method for patch classification. We also compare our methods on raw pixels of a patch. From the Fig.~\ref{roc_curve}, we can see that utilization of resampling features boosts the performance in patch classification. We also show our classification result in Table~\ref{tab:lstm}.
The results in Table~\ref{tab:lstm} and Fig.~\ref{roc_curve} attest that the method is effective in classifying tampered patches.

\begin{table}[H]
\centering
\begin{tabular}{|c|c|c|}
\hline
  Method & Classification & AUC\\ \hline
  Proposed Method & 94.86$\%$ & 0.9138 \\ \hline
  \end{tabular}
\caption{Results of patch classification for LSTM based approach}
\label{tab:lstm}
\end{table}

\section{Conclusion}
\label{sec:conc}

In this paper, we presented two methods to detect and localize manipulated regions in images.
Our experiments showed that both Convolutional Neural Networks (CNNs) and LSTM based networks are effective in exploiting resampling features to detect tampered regions.
In future, we will look into ways of combining these methods and detect image forgeries.

\section{Acknowledgements}
This research was developed with funding from the Defense Advanced Research Projects Agency (DARPA). The views, opinions and/or findings expressed are those of the author and should not be interpreted as representing the official views or policies of the Department of Defense or the U.S. Government.


{\small
\bibliographystyle{ieee}
\bibliography{ucr,forensics-mc,egbib}
}

\end{document}